\documentclass[11pt,a4paper]{article}
\usepackage[hyperref]{emnlp2020}
\usepackage{times}
\usepackage{graphicx}
\usepackage{latexsym}

\usepackage{microtype}

\aclfinalcopy 
\def\aclpaperid{19} 

\title{When Does Unsupervised Machine Translation Work?}

\author{Kelly Marchisio \and Kevin Duh \and Philipp Koehn \\
        Johns Hopkins University \\ {\tt kmarc@jhu.edu, kevinduh@cs.jhu.edu, phi@jhu.edu}}

\date{}

\begin{document}
\maketitle
\begin{abstract}
Despite the reported success of unsupervised machine translation, the field has yet to examine the conditions under which the methods succeed and fail. We conduct an extensive empirical evaluation using dissimilar language pairs, dissimilar domains, and diverse datasets.  We find that performance rapidly deteriorates when source and target corpora are from different domains, and that stochasticity during embedding training can dramatically affect downstream results.  We advocate for extensive empirical evaluation of unsupervised MT systems to highlight failure points and encourage continued research on the most promising paradigms. Towards this goal, we release our preprocessed dataset to stress-test systems under multiple data conditions. 
\end{abstract}

\section{Introduction}
Machine translation (MT) has progressed rapidly since the advent of neural machine translation (NMT) \citep{kalchbrenner2013recurrent,bahdanau2015neural,sutskever2014sequence} and is better than ever for languages for which ample high-quality bitext exists. Conversely, MT for low-resource languages remains a great challenge due to a dearth of parallel training corpora and poor quality bitext from esoteric domains.
To address this, several authors have proposed unsupervised MT techniques, which rely only on monolingual text for training \cite[e.g.,][]{ravi2011deciphering,yang-etal-2018-unsupervised,artetxe2018iclr,hoshen-wolf-2018-non,DBLP:conf/iclr/LampleCDR18,lample-phrase-2018,artetxe2018,artetxe2019}. 

Recent unsupervised MT results appear promising, but they primarily report results for the high-resource languages for which traditional MT already works well. The limits of these methods are so far under-explored. For unsupervised MT to be a viable path for low-resource machine translation, the field must determine (1) if it works outside highly-controlled environments, and (2) how to effectively evaluate newly-proposed training paradigms to pursue those which are promising for real-world low-resource scenarios. Unsupervised MT methods must work (1) on \textbf{different scripts} and between \textbf{dissimilar languages}, (2) with \textbf{imperfect domain alignment} between source and target corpora, (3) with a \textbf{domain mismatch} between training data and the test set, and (4) on the low-quality data of \textbf{real low-resource languages}. These factors reflect the real-life challenges of low-resource translation.

Our main contribution is an extensive analysis of unsupervised MT with regards to factors (1)-(3) above.\footnote{We release our full dataset at \url{http://statmt.org/when-does-unsup-work} to facilitate the stress-testing of systems.} We find that (a) translation performance rapidly deteriorates when source and target corpora are from different domains, (b) stochasticity during word embedding training can dramatically affect downstream bilingual lexicon induction (BLI) and translation performance, and (c) like in the bilingual lexicon induction literature, unsupervised MT performance declines when source and target languages are dissimilar. While (4) is not the focus of this paper, we do observe very low performance on an authentic low-resource language pair, corroborating previous studies \cite{guzman2019flores}.

Finally, as there are no standard evaluation protocols to ensure that unsupervised MT systems are robust to the types of data anomalies ubiquitous in low-resource translation settings, we advocate for extensive empirical evaluation of unsupervised MT systems to highlight failure points and encourage continued research on the most promising paradigms.

We first discuss related work in Section \ref{sec:related}, followed by a detailed overview of the unsupervised MT architecture in Section \ref{sec:background}. In Section \ref{sec:questions}, we discuss our research questions, followed by our evaluation methodology and datasets in Sections \ref{sec:protocol} and \ref{sec:data}. Section \ref{sec:reinvestigation} presents our findings, and Section \ref{sec:discussion} discusses the results. We conclude in Section \ref{sec:conclusion}.

\section{Related Work}
\label{sec:related}
\paragraph{Bilingual Lexicon Induction}Unsupervised MT methods can be thought of as an end-to-end extension of work inducing bilingual lexicons from monolingual corpora. Bilingual lexicon induction (BLI) using non-parallel data has a rich history, beginning with corpus statistic and decipherment methods \cite[e.g.,][]{rapp1995,fung-1995-compiling,koehn-knight-2000, koehn-knight-2002-learning, haghighi-etal-2008-learning}, continuing to modern neural methods to create crosslingual word embeddings \citep[e.g.][see \citet{ruder2019survey} for a survey]{mikolov2013exploiting, conneau-lample-2018} which form a critical component of state-of-the-art unsupervised MT systems.

\paragraph{Evaluation of Embedding Spaces} 
\citet{sogaard-etal-2018-limitations} determine that monolingual embedding spaces of similar languages are not typically isomorphic as was previously believed, and that bilingual dictionary induction ``depends heavily on... the language pair, the comparability of the monolingual corpora, and the parameters of the word embedding algorithms." \citet{vulic-etal-2019-really} argue that unsupervised approaches are unsuccessful with dissimilar languages and domains, and that unsupervised performance has been overly lauded because the conditions under which they were compared with supervised baselines were inequitable. 

While a modest body of literature has examined the quality of cross-lingual word embeddings (CLEs) by measuring performance on BLI, \citet{glavas-etal-2019-properly} evaluate on downstream natural language tasks, underlining the importance of full-system evaluation. The authors conclude that ``the quality of CLE models is largely task-dependent and that overfitting the models to the BLI task can result in deteriorated performance in downstream tasks." Similarly, \citet{doval2019robustness} investigate cross-lingual natural language inference.   

\paragraph{Evaluation of Unsupervised MT}
\citet{liu2020multilingual} helpfully re-define unsupervised machine translation into three distinct categories: (1) no bitext whatsoever, (2) the target language pair is linked through bitext via a pivot language, and (3) no linkage through a pivot language, but bitexts exists for *some* language and the target language. The authors analyze their multilingual pretraining method with respect to other similar training paradigms \citep{conneaulample2019cross, song2019mass} and evaluate unsupervised MT performance when using backtranslation (Definition 1) or language transfer after finetuning on related bitext (Definition 3).

In unsupervised MT with no bitext, \citet{lample-phrase-2018} ablate their PBSMT system, finding that initial phrase table quality is critical and that performance suffers when the language model is trained with less data. They tweak their NMT embedding initialization method, such as using separately-trained BPE instead of joint, and word embeddings instead of BPE token embeddings. They report the results of dropping part of their loss function and making minor changes to the NMT architecture on downstream BLEU score. Concurrently to our work, \citet{kim-unsup-useless-2020} arrived at similar conclusions to us using a different autoencoder/dual-learning unsupervised MT approach based on cross-lingual language model pretraining \citep{conneaulample2019cross}; this complements our experiments and corroborates our results.

\section{Background: Unsupervised MT}
\label{sec:background}

Our experiments employ the models of \citet{artetxe2018,artetxe2019} as representative of state-of-the-art for the class of unsupervised MT methods that bootstrap from cross-lingual word embeddings. Recent work such as \citet{lample-phrase-2018} is based on similar concepts. For our purposes, unsupervised MT follows \citet{liu2020multilingual}'s Definition (1) from Section \ref{sec:related}, where no bitext exists.

Another approach to unsupervised MT involves pretraining a bilingual or multilingual model on monolingual text on a general task before finetuning on translation. Such methods include cross-lingual language model pretraining \citep{conneaulample2019cross}, masked sequence-to-sequence pretraining \citep{song2019mass}, and multilingual denoising pretraining \citep{liu2020multilingual}, and have shown promise. For instance, \citet{liu2020multilingual} record the first good results on the low-resource Sinhala-English and Nepali-English pairs. While pretraining and multilingual methods are not the subject of this work, they warrant future evaluation.

\begin{figure}[thb]
  \centering
  \includegraphics[height=0.22\textheight,width=0.85\linewidth]{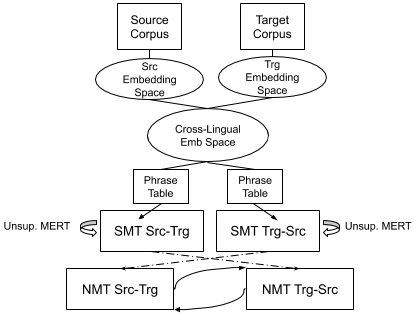}
  \caption{The unsupervised MT architecture used in this work. This model is a replication of \citet{artetxe2018} [steps before NMT] and \citet{artetxe2019} [NMT component].}
  \label{fig:artetxe_combined_arch}
\end{figure}

Figure~\ref{fig:artetxe_combined_arch} depicts the basic training process. It is the publicly-available SMT setup of \citet{artetxe2018}\footnote{\url{https://github.com/artetxem/monoses}}, plus the ``NMT hybridization" steps from \citet{artetxe2019}.\footnote{Shared with us by Mikel Artetxe.}

Training begins with two monolingual corpora which are not necessarily related in any way (i.e. they are not assumed to be parallel nor comparable text). 
First, word embeddings are trained independently for each corpus, resulting in a source and a target embedding space. 
Specifically, after preprocessing, \citet{artetxe2018} train two statistical language models using KenLM \cite{heafield-2011-kenlm}, one for the source language and one for the target. They use phrase2vec\footnote{\url{https://github.com/artetxem/phrase2vec}} \cite{artetxe2018}, an extension of \citet{mikolov2013}'s skip-gram model,\footnote{\url{https://github.com/tmikolov/word2vec}} to generate phrase embeddings for 200,000 unigrams, 400,000 bigrams, and 400,000 trigrams. 

Next, source and target word embeddings are aligned into a common cross-lingual embedding space. They run VecMap\footnote{\url{https://github.com/artetxem/vecmap}} \cite{artetxe-etal-2018-robust} which calculates a linear mapping of one space to another based on the intuition that phrases with similar meaning should have similar neighbors regardless of language. Given a matrix of source word embeddings $X$ and target word embeddings $Z$ which have been length-normalized, mean-centered, then length-normalized again, VecMap calculates $M_x=XX^T$ and $M_z=ZZ^T$. Each cell $M_{x_{ij}}$ and $M_{z_{ij}}$ is the cosine similarity between words $X_i$ and $X_j$, and $Z_i$ and $Z_j$, respectively. $M_x$ and $M_z$ are symmetric, and if the monolingual vector spaces were fully isometric, $M_x$ and $M_z$ would be identical besides rows and columns being permuted. Each row of $M_x$ and $M_z$ is a similarity distribution. To exploit this, each row of $\sqrt{M_x}$ and $\sqrt{M_z}$ is sorted (they find that using the square root works better empirically), and length-normalized, mean-centered, and length-normalized again. For each row $i$ in $sorted(\sqrt{M_x})$, they find the row $j$ of $sorted(\sqrt{M_z})$ that is its nearest neighbor, and assign $X_i=Z_j$ in the initial translation dictionary $D$. A cell $D_{ij} = 1$ if words $X_i$, and $Z_j$ are translations of one another, and 0 otherwise.

Next, there is an iterative process of calculating the optimal linear mappings and extracting an updated dictionary. For calculating the mapping, the goal is to find the linear transformations $W_x$ and $W_z$ which maximize the cosine similarity of the words that are translations of one another as defined by the dictionary $D$, over the entire dictionary:
\[
\arg\max_{W_x,W_z}\sum\limits_i\sum\limits_j (D_{ij})((X_iW_x) \cdot (Z_jW_z))
\]
From there, they calculate $M = XW_xW_z^TZ^T$, whereby each cell in $M$ is the cosine similarity of word $X_i$ and $Z_j$ after their transformations with $W_x$ and $W_z$. To avoid poor local optima, they stochastically zero-out some cells of $M$ with probability $p=0.9$, decreasing over time. 

The final score for each potential translation candidate is calculated using Cross-domain Similarity Local Scaling (CSLS) \cite{conneau-lample-2018} to mitigate the hubness problem. CSLS utilizes cosine similarity, which is taken from $M$. For each pair of words $X_i$ and $Z_j$, the new dictionary cell $D_{ij} = 1$ if the CSLS score between $X_i$ and $Z_j$ is the highest over all other words in $Z$, and $D_{ij} = 0$ otherwise. The dictionary is created in both directions, and concatenated. Readers are directed to \citet{artetxe-etal-2018-robust} for further details.

The next step extracts an initial phrase-table for use in a SMT system. They use the softmax over the cosine similarity of the 100 nearest-neighbors of each source phrase embedding as the phrase translation probabilities. This is done in both directions:
\[
(f|e) = \frac{e^{(\cos(e, f)/\tau)}}{\sum_{f'}{e^{(\cos(e, f')/\tau)}}}
\]
For the target embedding with the highest cosine similarity, the phrases are aligned, and unigram translation probabilities are multiplied to become the lexical weighting. 

Combining the preliminary phrase table with a distortion penalty and language model produces the initial unsupervised phrase-based SMT system \cite{koehn-etal-2007-moses}. The SMT model weights are tuned using a variant of MERT \cite{och-2003-minimum} designed for unsupervised scenarios, which uses 10,000 parallel sentences generated via backtranslation \cite{sennrich2016improving}. The SMT model then undergoes three rounds of iterative backtranslation. 

\citet{artetxe2019} extend their 2018 work by adding a critical ``NMT hybridization" final step, which achieves significant gains over SMT alone.\footnote{Readers are directed to \citet{artetxe2019} for additional changes that resulted in sizable BLEU \cite{papineni2002bleu} gains before the NMT phase.} An NMT system is trained using backtranslated output from SMT for one epoch. On the next epoch, a small number of sentences are backtranslated with the newly-trained NMT system and concatenated with a slightly smaller fraction of SMT-generated bitext. The procedure continues for 30 epochs, gradually increasing the percentage of synthetic training data created by the NMT system until all of the training data is NMT-generated. The NMT system is trained for an additional 30 epochs of iterative backtranslation using data generated fully by the NMT system of the previous epoch. The test set is translated with beam search using an ensemble of models saved at every tenth epoch (six total), resulting in BLEU scores of 33.2 and 26.4 (SacreBLEU \cite{post-2018-call}) on newstest2014 for French-English and German-English, respectively. 

We run \citet{artetxe2018,artetxe2019}'s implementation for our experiments. 
Specifically, neural models are Transformer-big \cite{vaswani_2017} trained with fairseq \cite{fairseq} on one NVIDIA GeForce GTX 1080Ti GPU. Models use shared embeddings, the Adam optimizer with $\beta_1$ = 0.9, $\beta_2$ = 0.98 \cite{kingma2014adam}, label smoothing, initial learning rate of 1e-07 warming up for 4000 steps to 5e-04 before decaying, and dropout \cite{srivastava2014dropout} probability of 0.3. We set optimizer delay to 4 to simulate 4 GPUs.

To elucidate the performance gap due to the unsupervised architecture, we build a standard supervised NMT system using the same neural architecture described above. We train until performance on the development set ceases to improve for 10 epochs. To parallel the unsupervised setup, we translate the test set using an ensemble of 6 models; We perform ensemble selection by performance on a validation set, selecting the best-performing checkpoint along with 5 previous checkpoints.

\section{Research Questions} 
\label{sec:questions}

Existing unsupervised translation methods work well on languages which are similar to each other, use the same Roman script, and have an ample amount of monolingual news data available (which matches the test set domain). Questions remain as to whether unsupervised methods will be useful on authentic low-resource settings where few or none of the aforementioned properties hold. Namely, does unsupervised MT work with:
\begin{itemize}
\item dissimilar languages?
\item dissimilar source and target domains?
\item diverse datasets?
\item authentic low-resource language pairs?
\end{itemize}

Such questions reflect the reality of authentic low-resource translation, and are those which must be adequately resolved for unsupervised MT to be a viable alternative to traditional translation methods for the most difficult language pairs. 

\section{Evaluation of Unsupervised MT}
\label{sec:protocol}
We perform an extensive empirical evaluation of unsupervised MT. Our evaluation protocol stress-tests an unsupervised MT system under varying conditions to reveal its points of strength and failure. Systems should be judged on how well they perform: (1) on dissimilar languages, (2) on increasingly divergent domains between source and target corpora, (3) on diverse datasets, and (4) on authentic low-resource language pairs where data quality is typically low.  Namely, we:

\begin{enumerate}
\item Choose 2 language pairs, at least one of which where the source and target languages utilize different scripts.
\item Choose 3 datasets of different domains, at least one of which is parallel bitext. 
\item Perform at least one experiment for each language pair under each of the following data conditions:
\begin{itemize}
\item Originally parallel
\item Not originally parallel
\item Different domain for source and target.
\end{itemize}
\item Choose 2 \textit{true} low-resource language pairs.
\item Judge the system based on performance in all tested scenarios. 
\end{enumerate}

The data conditions above are designed measure how well a system performs in regards to the research questions of Section~\ref{sec:questions}. Namely, success on a variety of languages with different scripts and linguistic structure indicates robustness to dissimilar languages; success on multiple datasets of different domains indicates that the system is not specifically designed for one domain at the expense of others, and performs well even when training and test data do not match perfectly; Step \#3 evaluates performance on increasingly divergent domains between source and target data; and Step \#4 is the \textit{true} test---whether the system succeeds on authentic low-resource language pairs.

\section{Datasets}
\label{sec:data}
Training datasets used in our reinvestigation of the unsupervised MT system presented in \citet{artetxe2019} are shown in Table \ref{data}.
We focus on Russian-English (Ru-En) and French-English (Fr-En) tasks and include as reference Sinhala-English (Si-En) and Nepali-English (Ne-En) as well. 
Following Section \ref{sec:protocol}, we evaluate the same system under various ablated data setups:
\begin{itemize}
\item The ``Supervised" condition is the standard MT training setup which uses parallel bitext.
\item 
In the ``Parallel" condition, an \textit{unsupervised} MT system is trained on a corpus that was originally parallel (i.e. UN corpus), now being treated as two separate monolingual corpora. 
\item In contrast, the ``Disjoint" setting splits data from a parallel corpus into two disjoint halves, using the first half of the source-side corpus and the second half of the target-side corpus. 
\item In the ``Different Domain" (Diff. Dom.) setting, source and target monolingual corpora come from different domains. This is a realistic setting in low-resource scenarios, and is expected to be much more difficult than the ``Disjoint" setting. 
\item ``News crawl" (News) and ``Common Crawl" (CC) settings determine whether the system can flexibly handle diverse datasets. 
\end{itemize}
Specifics of the datasets used are described in subsequent subsections. Token counts presented in the subsections below are before preprocessing, whereas Table \ref{data} reflects the data remaining after the preprocessing procedure of \citet{artetxe2018}. We will release the preprocessed data splits for others to compare their results with ours.

\begin{table}[t]
\centering
\setlength\tabcolsep{4pt} 
\begin{tabular}{l l|l|l|l}
\hline \textbf{Condition} & & \textbf{Corpus} & \textbf{Src} & \textbf{Trg} \\ \hline
Repro & Fr-En & News & 694 & 1940 \\
& En-Fr & News & 1940 & 694 \\
Supervised & Fr-En & UN: A & 346 & 301  \\
& Ru-En & UN: A & 284 & 284 \\
Parallel & Fr-En & UN: A & 302 & 270 \\ 
& Ru-En & UN: A & 232 & 241  \\ 
Disjoint & Fr-En & UN: A / B & 302 & 255 \\ 
& Ru-En & UN: A / B & 232 & 236 \\ 
Diff.~Dom. & Fr-En & UN: A / CC & 302 & 226 \\ 
& Ru-En & UN: A / CC & 232 & 226 \\ 
News & Fr-En & News & 116 & 105  \\ 
& Ru-En & News & 120 &  105 \\ 
CC & Fr-En & CC & 110 & 79 \\ 
& Ru-En & CC & 115 & 79 \\ 
\hline
\end{tabular}
\caption{\label{data} Training data after preprocessing. UN = United Nations, CC = Common Crawl, News = News crawl. ``Diff. Dom." uses UN on the source-side and CC on the target-side. ``News" is a subset of 2007-08 for En, 2007-09 for Fr, and 2008-11 for Ru. ``Repro" is the condition most similar to \cite{artetxe2018,artetxe2019}. Src (M) and Trg (M) columns are the token counts, in millions. ``Supervised" count is in BPE tokens. All others are token counts for SMT (pre-BPE).}
\end{table}

\subsection{United Nations}
The United Nations Parallel Corpus (UN) \cite{ziemski-etal-2016-united} contains official United Nations documents from 1990-2014, human-translated into six languages. The first 10,000 lines of each dataset are held-out. The remaining lines are partitioned into training sets A \& B. Training set A on the source side and A on the target side are paired to form the Parallel training set; Training set A on the source side and B on the target side are paired to form the Disjoint training set.

\subsection{News Crawl}\label{news}
News crawl (News) consists of monolingual data crawled from news websites. Data for each year has been shuffled. Following \citet{artetxe2018}, we concatenate News crawl 2007-13 for English and for French. For Russian, we concatenate News crawl 2008-18.  We use the deduplicated Russian corpus. We use the full datasets to reproduce \citet{artetxe2018, artetxe2019}'s work. For subsequent experiments, we use a subset: the first 100 million tokens from each concatenated News crawl corpus before preprocessing. For English, this is all of News crawl 2007 and $\sim$23.3 million tokens from News crawl 2008.  For French, it is News crawl 2007, 2008, and some of 2009. For Russian, it is News crawl 2008-2010, and some of 2011. 

\subsection{Common Crawl}
The Common Crawl (CC) corpora consists of web-scraped monolingual data ordered as documents. We extract two training datasets from the English corpus - one with the first $\sim$291 million tokens and another with the first $\sim$100 million for Diff. Dom and CC experiments, respectively.  We do not shuffle this data, as having less documents better simulates real low-resource settings. Sinhala and Nepali contain approximately 103 million and 110 million tokens, as used in \citet{guzman2019flores}. We additionally extract the first 100 million French and Russian tokens for CC experiments.

\subsection{Preprocessing}\label{preprocess}
Training data is preprocessed separately for each unsupervised experiment as part of \citet{artetxe2018}'s training pipeline. Data is deduplicated, and tokenized and truecased using scripts from Moses \cite{koehn-etal-2007-moses}. Sentences with less than 3 tokens or more than 80 tokens are discarded, and sentences are shuffled. Ten thousand sentences are removed to form a development set.  To begin the NMT phase, a joint BPE \cite{sennrich-etal-2016-neural} vocabulary of 32000 tokens is learned. Source- and target-side corpora are backtranslated using the final model from the SMT phase, and all data then has BPE applied.\footnote{Some experiments had Moses' \url{clean-corpus-n.perl} applied after this.}

For supervised experiments, training data is tokenized and truecased, and then a joint BPE \cite{sennrich-etal-2016-neural} vocabulary of 32000 tokens is learned. After applying BPE, the data is cleaned using Moses' \texttt{clean-corpus-n.perl}, discarding sentences under 3 and greater than 80 tokens.

\subsection{Vocabulary Overlap of Training Sets}
A vocabulary of unigrams was collected for each target-side (English) corpus, which includes tokens that appear at least 10 times, for a maximum of 200,000 unigrams.  Of approximately 144,000 such unique tokens between UN-A and UN-B from the Fr-En UN corpus, the corpora share 54.1\%. These corpora are used in the Disjoint condition. The respective vocabulary overlap for UN-A and CC from the Diff.~Dom condition for Fr-En is 25.7\%. For UN-B vs. CC for Fr-En, they share 25.3\%. Statistics are analogous for Ru-En.

\subsection{Test and Validation Sets}
Ru-En models are tested on newstest2019. Fr-En models are tested on newstest2014. Supervised models use newstest2018 (Ru-En) or newstest2013 (Fr-En) for validation. For Si-En and Ne-En, we use the Wikipedia dev and devtest sets from \citet{guzman2019flores}.\footnote{\url{https://github.com/facebookresearch/flores/raw/master/data/wikipedia\_en\_ne\_si\_test\_sets.tgz}} For supervised models, we select the ensemble with best performance on newstest2017 (Ru-En) or newstest2012 (Fr-En).

\section{Reinvestigation of Artetxe et. al.}

First, we replicate \citet{artetxe2018, artetxe2019}, achieving relatively comparable results (Table~\ref{tab:repro_results}). Differences in BLEU scores are likely attributable to using \citet{artetxe2018}'s code for all steps before the NMT phase; \citet{artetxe2019} improved upon these, but we chose to use the publicly available code from the previous year. 

\label{sec:reinvestigation}
\begin{table}[h]
\centering
\begin{tabular}{l l l l}
\hline & \textbf{\citet{artetxe2019}} & \textbf{This Work}\\ \hline
Fr-En&33.2&31.1& \\
En-Fr&33.6&32.8& \\
\end{tabular}
\caption{\citet{artetxe2019}'s unsupervised MT performance vs. the system in this work, which is a combination of \citet{artetxe2018} [steps before NMT] and \citet{artetxe2019} [NMT component], using the full News crawl datasets from Subsection \ref{news}. Scored using SacreBLEU \cite{post-2018-call} on newstest2014.}
\label{tab:repro_results}
\end{table}

Next, we set up a series of experiments to assess the questions posed in Section \ref{sec:questions}. Results are presented in Tables \ref{tab:un_results} and \ref{tab:other_dataset_results}.

\subsection{Unsupervised Quality Loss}

The Supervised (``Sup.") column of Table \ref{tab:un_results} shows performance of a standard Transformer-big architecture on parallel bitext for Ru-En and Fr-En. Assuming that supervised translation will always outperform unsupervised, these scores represent a ceiling to quantify how much potential quality is lost using an unsupervised architecture. 

\begin{table}[h]
\centering
\setlength\tabcolsep{3pt} 
\begin{tabular}{l l l l l}
\hline & \textbf{Sup.} & \textbf{Parallel} & \textbf{Disjoint} & \textbf{Diff.~Dom.}\\ 
\textit{Corpus} & A / A & A / A & A / B & A / CC*\\ \hline
Ru-En & 26.9 
& 23.7 \textcolor{red}{\small{\textit{(-3.2)}}} 
& 21.2 \textcolor{red}{\small{\textit{(-5.7)}}}
& 0.7 \textcolor{red}{\small{\textit{(-26.2)}}} \\
Fr-En & 29.9 
& 27.6 \textcolor{red}{\small{\textit{(-2.3)}}}
& 27.0 \textcolor{red}{\small{\textit{(-2.9)}}}
& 3.9 \textcolor{red}{\small{\textit{(-26.0)}}} \\
\end{tabular}
\caption{Unsupervised MT performance on a single run using the United Nations (UN) dataset.  ``Diff. Dom." uses UN data as source and Common Crawl (*) as target.  ``Sup." is supervised with UN parallel data. A / A refers to UN training dataset A used on the source and target sides, for example. Scored using SacreBLEU \cite{post-2018-call} on newstest2019 (Ru-En) and newstest2014 (Fr-En).}
\label{tab:un_results}. 
\end{table}

The supervised models and those in the Parallel column use the same datasets\footnote{Differences in token count are due to the different preprocessing detailed in Section \ref{preprocess}.} and can therefore be directly compared.  We observe a BLEU score drop of $\sim$3.2 for Ru-En versus a drop of $\sim$2.3 for Fr-En when using the unsupervised architecture. 
This minor quality loss represents a strong result for unsupervised MT; however, the question is whether the results will remain strong as we gradually make the monolingual corpora less similar.

\subsection{Investigating Our Research Questions}
\textit{Does unsupervised machine translation work for:}

\vspace{1mm}\noindent\textit{(1) Dissimilar language pairs?}

We conduct experiments in French and Russian into English. Whereas French and English share the same Roman script and common linguistic origin, Russian is a Slavic language that uses the Cyrillic script. The results presented in Tables \ref{tab:un_results} and \ref{tab:other_dataset_results} indicate that unsupervised MT is more difficult when writing script and language family differs. Across the board, we observe that the $\Delta$BLEU between supervised and unsupervised performance is wider for Ru-En than for Fr-En, particularly for News and Common Crawl datasets. For instance, whereas Fr-En loses 2.9 BLEU in the Supervised versus Disjoint setups (which use comparable data), Ru-En loses 5.7 BLEU. While we acknowledge that in general one should not compare BLEU scores across language pairs or datasets, this gap suggests that unsupervised MT may behave differently for different language pairs.

\vspace{2mm}\noindent\textit{(2) Dissimilar domains?}

We investigate the effects of domain similarity between source and target training corpora. For each language, we observe the difference in performance on Table \ref{tab:un_results} of the Parallel, Disjoint, and Diff. Dom. columns.

Because training data in the Parallel condition was originally parallel, these experiments have the highest possible domain match between source and target data. Since Disjoint data was extracted from the same corpus but was not parallel, source and target can be thought of as having very slightly different domains. We observe a minor performance drop between Parallel and Disjoint experiments, which is more pronounced for Ru-En. 

Examining the Diff. Dom. column, however, the performance contrast is stark. While both language pairs obtain respectable BLEU scores in the 20s when domains match in Parallel and Disjoint conditions, performance drops sharply when training set domains are mismatched---scoring 3.9 BLEU for Fr-En and 0.7 for Ru-En. (A subsequent run of Fr-En scored 17.4, addressed in Section \ref{stability}). The fault is not with either side of the training corpus alone---Parallel/Disjoint experiments from Table \ref{tab:un_results} which use UN data alone and CC experiments in Table \ref{tab:other_dataset_results} which use Common Crawl data alone perform acceptably---it is when the two datasets are paired as source-target in Diff. Dom. conditions that performance rapidly deteriorates.

\vspace{2mm}\noindent\textit{(3) Diverse datasets?}

\begin{table}[h]
\centering
\begin{tabular}{l l l l}
\hline & \textbf{UN} & \textbf{News} & \textbf{CC}\\ \hline
Ru-En & 21.2 & 16.1 & 13.8  \\
Fr-En & 27.0 & 28.2 & 22.4  \\
Si-En & n/a & n/a & 0.2 \\
Ne-En & n/a & n/a & 0.4 \\
\end{tabular}
\caption{Unsupervised MT performance on a single run using diverse datasets [UN = United Nations (Disjoint), News = News Crawl, CC = Common Crawl]. Scored using SacreBLEU \cite{post-2018-call} on newstest2019 (Ru-En), newstest2014 (Fr-En), and the FLoRes Wikipedia evaluation sets (Si-En, Ne-En) \cite{guzman2019flores}.}
\label{tab:other_dataset_results}
\end{table}

Table \ref{tab:other_dataset_results} shows the results of experiments using three different training datasets. News crawl matches the domain of the test set exactly. UN data has a moderate domain match with the test set, and CC matches the least. Not unexpectedly, most experiments where training and test domain match perform better than when there is a domain mismatch. The exception is the News experiment for Ru-En, where the model performs considerably worse than the UN condition despite having a stronger domain match. Notably, News has approximately 2-3x less data than UN for each language pair. We suspect that for Fr-En, the relative ease of unsupervised translation for this language pair allowed the strong domain match with the test set to outweigh the lower amount of data. On the other hand, the relative difficulty of unsupervised MT in Ru-En made the system suffer too greatly in the lower-resource condition, to where it could not compensate with domain match. 

\vspace{2mm}\noindent\textit{(4) A true low-resource pair?}

Facebook recently released test sets for Sinhala-English and Nepali-English, true low-resource language pairs which not only lack bitext, but monolingual data is of poor quality. These languages do not share a script or language family with English, and the data is out-of-domain with the English data. This reflects a real-world low-resource scenario where we would hope to benefit from unsupervised MT. We observe extremely poor results in Table \ref{tab:other_dataset_results}, with Si-En achieving a BLEU score of 0.2, and 0.4 for Ne-En. \citet{guzman2019flores} achieve similarly poor results for these language pairs without using supplemental data from a related language. 

\subsection{BLEU During Training}
\begin{figure}[h]
  \centering
  \includegraphics[height=0.20\textheight,width=0.90\linewidth]{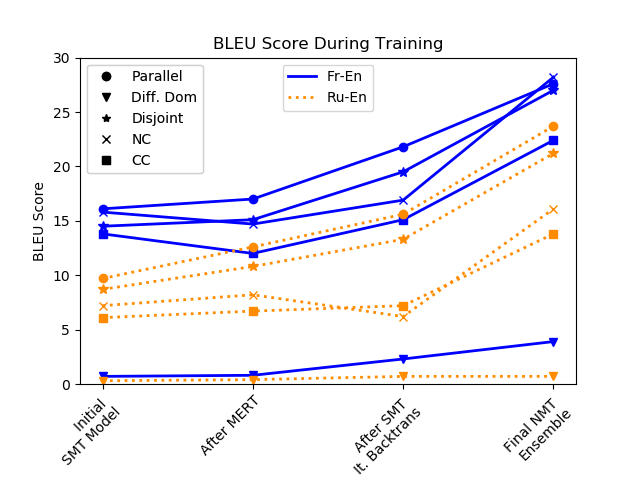}
  \caption{BLEU score during training.}
  \label{fig:bleu_during_training}
\end{figure}

Figure \ref{fig:bleu_during_training} shows translation performance for the experiments in Tables \ref{tab:un_results} and \ref{tab:other_dataset_results} at various steps during the unsupervised machine translation pipeline. Most SMT models improve performance slightly as a result of unsupervised MERT tuning, and more substantially after three rounds of iterative backtranslation. Substantial improvement occurs as a result of NMT training for all models except the degenerate Diff.~Dom conditions. 

\subsection{Training Stability}\label{stability}

One challenge with unsupervised methods is training stability: stochasticity during training can give substantially different results due to the iterative bootstrap nature of the training process.

In their analysis of unsupervised methods for generating CLEs, \citet{glavas-etal-2019-properly} note considerable instability in performance on BLI. Defining failure as having a mean average precision (MAP) of $<$0.05 on all training runs, Iterative Closest Point \cite{hoshen-wolf-2018-non} fails for $\sim$21\% of language pairs, Gromov-Wasserstein Alignment \cite{alvarez-melis-jaakkola-2018-gromov} for $\sim$46\%, and MUSE \cite{conneau-lample-2018} for $\sim$54\%. VecMap \cite{artetxe-etal-2018-robust} succeeds for all language pairs, leading \citeauthor{glavas-etal-2019-properly} to deem it the most robust. \citet{artetxe-etal-2018-robust} demonstrate their robustness over other methods in their work. When counting successful runs as achieving $>$5.0\% accuracy, VecMap is successful 10/10 times for three language pairs. \citet{hartmann_NIPS2019_8836} also investigate instability in vector space alignment methods.

\begin{table}
\centering
\setlength\tabcolsep{3.5pt} 
\begin{tabular}{l l|l|l|l|l}
\hline & \textbf{Condition} & \textbf{Min} & \textbf{Max} & \textbf{$\mu$} & \textbf{$\sigma$}\\ \hline
En-Fr & Repro &  33.08  &  42.47  &  40.86  &  2.5  \\
Fr-En & Repro &  45.21  &  46.92  &  46.06  &  0.47  \\
    & Parallel &  48.0  &  50.2  &  49.09  &  0.69  \\
    & Disjoint &  37.88  &  39.09  &  38.47  &  0.37  \\
    & Diff. Dom. &  \textbf{0.0}  &  \textbf{17.27}  &  \textbf{7.97}  &  \textbf{7.95}  \\
    & News &  25.86  &  28.1  &  26.97  &  0.56  \\
    & CC &  25.87  &  27.6  &  26.9  &  0.51  \\
Ru-En & Parallel &  32.24  &  34.04  &  32.95  &  0.47  \\
    & Disjoint & 25.08  &  26.96  &  25.79  &  0.58  \\
    & Diff. Dom. &  0.0  &  0.1  &  0.01  &  0.03  \\
    & News &  22.19  &  23.77  &  23.1  &  0.44  \\
    & CC &  \textbf{0.0}  &  \textbf{24.69}  &  \textbf{12.61}  &  \textbf{11.45}  \\
\hline
\end{tabular}
\caption{\label{tab:instability} Accuracies (\%) of induced dictionaries on 10-11 runs. Bold experiments were severely unstable.}
\end{table}

After training phrase embeddings for each experiment, we run VecMap on the generated embedding spaces ten additional times and indeed find little fluctuation in BLI between runs. When rerunning the full pipeline for each experiment, however, we observe considerable instability in two experiments which dramatically affects downstream results. 

We build a gold-standard bilingual dictionary of 2000 word pairs from Wikipedia data \cite{wolk2014building} available publicly on OPUS \cite{tiedemann2012parallel}, and run the first four steps of the unsupervised training procedure additional times for each experiment. Table \ref{tab:instability} contains the summary results of 10-11 runs of each experiment.

Tables \ref{tab:un_results} and \ref{tab:other_dataset_results} present the results of the single first run of each experiment. Whereas the majority have consistent accuracy on bilingual lexicon between runs as seen in Table \ref{tab:instability}, Diff. Dom. for Fr-En and CC for Ru-En are highly unstable. The BLI accuracy of additional runs of Fr-En Diff. Dom. ranged between 0.0\% and 17.27\%. Of the initial run and 9 subsequent, five had accuracies $<$0.1\%, while the other five had accuracies $>$15.26\%. For Ru-En CC, the run reported in Table \ref{tab:other_dataset_results} had a BLI accuracy of 21.35\%. Of eleven runs, five had an accuracy $<$0.26\%, and six had an accuracy $>$21.35\%.

As evidence of the critical effect of BLI accuracy on downstream BLEU, whereas the Fr-En Diff. Dom. run reported in Table \ref{tab:un_results} had a BLI accuracy of 0.0\%, a subsequent run of the entire training pipeline had an accuracy of 17.08\% and a final BLEU score of 17.4. (This experiment is not included in the summary statistics of Table \ref{tab:instability}).

The unsupervised pipeline begins with preprocessing (deterministic, except shuffling and random selection of development set), language model training with KenLM \cite{heafield-2011-kenlm} (deterministic), followed by phrase embedding training using phrase2vec (non-deterministic), and then embedding space mapping with VecMap (non-deterministic).  Because performance on reruns of VecMap alone was stable while holding the rest of the system constant, we must conclude that the dramatic instability is caused by either a poor embedding initialization from phrase2vec/word2vec, or VecMap's inability to handle certain monolingual vector space configurations. Apparently, the initial formation of monolingual vector spaces dramatically affects VecMap's ability to converge to a good solution, which in turn results in highly variable downstream translation performance.

\begin{figure}[h]
  \centering
  \includegraphics[width=1.0\linewidth]{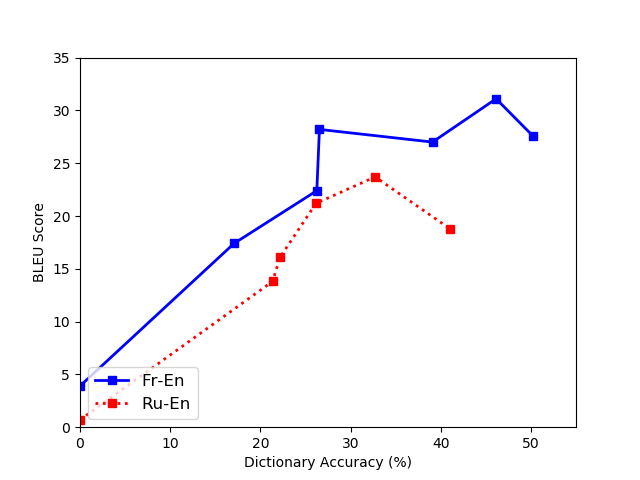}
  \caption{Relationship between bilingual lexicon induction accuracy after VecMap mapping, and final BLEU.}
  \label{fig:dicacc}
\end{figure}

To observe the relationship between BLI accuracy and downstream BLEU score, we direct the reader to Figure \ref{fig:dicacc}, where BLI accuracy after the VecMap phase of experiments from Tables \ref{tab:un_results} and \ref{tab:other_dataset_results} are displayed in relation to the final BLEU score.

\section{Discussion}
\label{sec:discussion}
Except in the Diff. Dom. condition, unsupervised MT performance for Fr-En is impressive and suggests that sentence alignment may not be required for successful MT under ideal conditions. Ru-En results are also impressive, but show that unsupervised MT still struggles when language pairs are dissimilar, especially when data amount is reduced. 

The gap between Disjoint and Diff. Dom. conditions is perhaps the most striking result in our experiments. It suggests that one cannot naively collect monolingual corpora without considering their relative domain similarity; this may be a challenge in low-resource conditions, where there is less flexibility with data sources. 
\citet{vulic-etal-2019-really} make a similar claim about unsupervised CLEs, stating ``UNSUPERVISED methods are able to yield a good solution only when there is no domain mismatch and for the pair with two most similar languages (English-Spanish), again questioning their robustness and portability to truly low-resource and more challenging setups". 
Furthermore, the extremely poor results of Ne-En and Si-En reflect the reality of low-resource translation; the compound negative effects of language dissimilarity, domain mismatch between monolingual corpora, domain mismatch with the test set, and low amounts of low-quality data. It is the ``worst of all worlds"---but reflects how current models might perform on the use cases for which they are needed. These challenges highlight the importance of evaluating unsupervised MT under varying realistic data conditions. Our evaluation is a step towards this goal, and identifies multiple areas for improvement. 

A critical step in state-of-the-art unsupervised MT is methods for creating CLEs. Several authors have pointed out that ``mapping" methods like VecMap assume that monolingual vector spaces are structurally similar, but that this ``approximate isomorphism assumption" is increasingly tenuous as languages and domains diverge \cite{sogaard-etal-2018-limitations,ormazabal-etal-2019-analyzing, glavas-etal-2019-properly, vulic-etal-2019-really, patra-etal-2019-bilingual}. \citet{patra-etal-2019-bilingual} find this for Fr-En and Ru-En specifically, the languages examined in this work. \citet{nakashole-flauger-2018-characterizing} argue that while linearity may hold within local ``neighborhoods" of the vector space, the global mapping is non-linear. \citet{sogaard-etal-2018-limitations} use their eigenvector similarity metric to show a strong correlation between vector space similarity and BLI performance.  Analysis of the CLEs from our experiments demonstrate a relationship between BLI performance and downstream BLEU on the translation task. Coupled with our empirical evidence, the works cited in this section suggest that nonisometric vector spaces lead to poor quality translation.

Factors observed in our experiments that lead to lower quality translation can be attributable to a ``weak isomorphism" between the monolingual vector spaces. Dissimilar languages means increasingly different distributional characteristics of words. Data from different domains naturally have different word frequencies and distributional characteristics, which become more pronounced as domains diverge. Because mapping methods rely on structural similarity of vector spaces, experiments using either UN or CC data alone had acceptable downstream performance, where as combining the datasets as source and target resulted in extremely poor translation.  We observe the critical effect of word embedding initialization on BLI performance and downstream BLEU, suggesting that stochasticity during word embedding creation can cause resulting vector spaces to be more or less isomorphic. Finally, more data can give a more accurate distribution of words in comparison with the true distribution in the language, leading to a more realistic monolingual vector space. With less data, word embeddings are dependent on the smaller training sample, which may not match the test set or reflect true distributional properties of the language. Combining all of these negative factors likely leads to highly nonisomorphic monolingual embedding spaces, as demonstrated by the very poor Si-En and Ne-En results.

\section{Conclusion \& Future Work}
\label{sec:conclusion}
Progress in unsupervised MT has been impressive, achieving performance near its supervised counterparts under some scenarios. That said, evaluating current approaches under broader settings and datasets reveals that unsupervised MT struggles in realistic low-resource scenarios.  
As stated by \citet{lample-phrase-2018}, ``It's an open question whether there are more effective instantiations of these principles [underlying recent successes in fully unsupervised MT] or other principles altogether". In this work, we find that there is room for improvement to become robust to (1) dissimilar languages pairs, (2) dissimilar domains, (3) diverse datasets, and (4) the low-quality data of true low-resource languages---factors ubiquitous in low-resource language pairs for which unsupervised MT is intended. We find that (a) performance rapidly declines when source and target corpora are from different domains, and (b) stochasticity during word embedding training can dramatically affect downstream translation results. The latter is a yet unexplored research area. Future work should also evaluate pretraining methods in bilingual and multilingual training contexts. 

Finally, we argue for extensive evaluation of unsupervised MT systems under varying data conditions to assess failure cases and encourage pursuit of promising paradigms. Doing so is a step towards solving the real-world problems of low-resource machine translation.

\section*{Acknowledgements}
The authors would like to thank Mikel Artetxe for providing an implementation of his 2019 paper and thoughtful feedback, Matthew Francis-Landau, Cheng-I (Jeff) Lai, and Huda Khayrallah for support and advice. We also thank our anonymous reviewers and colleagues at Johns Hopkins University for helpful feedback.

This material is based upon work supported by the United States Air Force under Contract No. FA8750‐19‐C‐0098.  Any opinions, findings, and conclusions or recommendations expressed in this material are those of the author(s) and do not necessarily reflect the views of the United States Air Force and DARPA.

\bibliographystyle{acl_natbib}
\bibliography{emnlp2020}

\begin{thebibliography}{49}
\expandafter\ifx\csname natexlab\endcsname\relax\def\natexlab#1{#1}\fi

\bibitem[{Alvarez-Melis and
  Jaakkola(2018)}]{alvarez-melis-jaakkola-2018-gromov}
David Alvarez-Melis and Tommi Jaakkola. 2018.
\newblock \href {https://doi.org/10.18653/v1/D18-1214} {{G}romov-{W}asserstein
  alignment of word embedding spaces}.
\newblock In \emph{Proceedings of the 2018 Conference on Empirical Methods in
  Natural Language Processing}, pages 1881--1890, Brussels, Belgium.
  Association for Computational Linguistics.

\bibitem[{Artetxe et~al.(2018{\natexlab{a}})Artetxe, Labaka, and
  Agirre}]{artetxe-etal-2018-robust}
Mikel Artetxe, Gorka Labaka, and Eneko Agirre. 2018{\natexlab{a}}.
\newblock \href {https://doi.org/10.18653/v1/P18-1073} {A robust self-learning
  method for fully unsupervised cross-lingual mappings of word embeddings}.
\newblock In \emph{Proceedings of the 56th Annual Meeting of the Association
  for Computational Linguistics (Volume 1: Long Papers)}, pages 789--798,
  Melbourne, Australia. Association for Computational Linguistics.

\bibitem[{Artetxe et~al.(2018{\natexlab{b}})Artetxe, Labaka, and
  Agirre}]{artetxe2018}
Mikel Artetxe, Gorka Labaka, and Eneko Agirre. 2018{\natexlab{b}}.
\newblock \href {https://doi.org/10.18653/v1/D18-1399} {Unsupervised
  statistical machine translation}.
\newblock In \emph{Proceedings of the 2018 Conference on Empirical Methods in
  Natural Language Processing}, pages 3632--3642, Brussels, Belgium.
  Association for Computational Linguistics.

\bibitem[{Artetxe et~al.(2019)Artetxe, Labaka, and Agirre}]{artetxe2019}
Mikel Artetxe, Gorka Labaka, and Eneko Agirre. 2019.
\newblock \href {https://doi.org/10.18653/v1/P19-1019} {An effective approach
  to unsupervised machine translation}.
\newblock In \emph{Proceedings of the 57th Annual Meeting of the Association
  for Computational Linguistics}, pages 194--203, Florence, Italy. Association
  for Computational Linguistics.

\bibitem[{Artetxe et~al.(2018{\natexlab{c}})Artetxe, Labaka, Agirre, and
  Cho}]{artetxe2018iclr}
Mikel Artetxe, Gorka Labaka, Eneko Agirre, and Kyunghyun Cho.
  2018{\natexlab{c}}.
\newblock Unsupervised neural machine translation.
\newblock In \emph{Proceedings of the Sixth International Conference on
  Learning Representations}.

\bibitem[{Bahdanau et~al.(2015)Bahdanau, Cho, and Bengio}]{bahdanau2015neural}
Dzmitry Bahdanau, Kyunghyun Cho, and Yoshua Bengio. 2015.
\newblock \href {http://arxiv.org/abs/1409.0473} {Neural machine translation by
  jointly learning to align and translate}.
\newblock In \emph{3rd International Conference on Learning Representations,
  {ICLR} 2015, San Diego, CA, USA, May 7-9, 2015, Conference Track
  Proceedings}.

\bibitem[{Conneau and Lample(2019)}]{conneaulample2019cross}
Alexis Conneau and Guillaume Lample. 2019.
\newblock Cross-lingual language model pretraining.
\newblock In \emph{Advances in Neural Information Processing Systems}, pages
  7057--7067.

\bibitem[{Conneau et~al.(2018)Conneau, Lample, Ranzato, Denoyer, and
  J{\'{e}}gou}]{conneau-lample-2018}
Alexis Conneau, Guillaume Lample, Marc'Aurelio Ranzato, Ludovic Denoyer, and
  Herv{\'{e}} J{\'{e}}gou. 2018.
\newblock \href {https://openreview.net/forum?id=H196sainb} {Word translation
  without parallel data}.
\newblock In \emph{6th International Conference on Learning Representations,
  {ICLR} 2018, Vancouver, BC, Canada, April 30 - May 3, 2018, Conference Track
  Proceedings}. OpenReview.net.

\bibitem[{Doval et~al.(2019)Doval, Camacho-Collados, Espinosa-Anke, and
  Schockaert}]{doval2019robustness}
Yerai Doval, Jose Camacho-Collados, Luis Espinosa-Anke, and Steven Schockaert.
  2019.
\newblock \href {https://arxiv.org/pdf/1908.07742.pdf} {On the robustness of
  unsupervised and semi-supervised cross-lingual word embedding learning}.
\newblock \emph{arXiv preprint arXiv:1908.07742}.

\bibitem[{Fung(1995)}]{fung-1995-compiling}
Pascale Fung. 1995.
\newblock \href {https://www.aclweb.org/anthology/W95-0114} {Compiling
  bilingual lexicon entries from a non-parallel {E}nglish-{C}hinese corpus}.
\newblock In \emph{Third Workshop on Very Large Corpora}.

\bibitem[{Glava{\v{s}} et~al.(2019)Glava{\v{s}}, Litschko, Ruder, and
  Vuli{\'c}}]{glavas-etal-2019-properly}
Goran Glava{\v{s}}, Robert Litschko, Sebastian Ruder, and Ivan Vuli{\'c}. 2019.
\newblock \href {https://doi.org/10.18653/v1/P19-1070} {How to (properly)
  evaluate cross-lingual word embeddings: On strong baselines, comparative
  analyses, and some misconceptions}.
\newblock In \emph{Proceedings of the 57th Annual Meeting of the Association
  for Computational Linguistics}, pages 710--721, Florence, Italy. Association
  for Computational Linguistics.

\bibitem[{Guzm{\'a}n et~al.(2019)Guzm{\'a}n, Chen, Ott, Pino, Lample, Koehn,
  Chaudhary, and Ranzato}]{guzman2019flores}
Francisco Guzm{\'a}n, Peng-Jen Chen, Myle Ott, Juan Pino, Guillaume Lample,
  Philipp Koehn, Vishrav Chaudhary, and Marc{'}Aurelio Ranzato. 2019.
\newblock \href {https://doi.org/10.18653/v1/D19-1632} {The {FLORES} evaluation
  datasets for low-resource machine translation: {N}epali{--}{E}nglish and
  {S}inhala{--}{E}nglish}.
\newblock In \emph{Proceedings of the 2019 Conference on Empirical Methods in
  Natural Language Processing and the 9th International Joint Conference on
  Natural Language Processing (EMNLP-IJCNLP)}, pages 6098--6111, Hong Kong,
  China. Association for Computational Linguistics.

\bibitem[{Haghighi et~al.(2008)Haghighi, Liang, Berg-Kirkpatrick, and
  Klein}]{haghighi-etal-2008-learning}
Aria Haghighi, Percy Liang, Taylor Berg-Kirkpatrick, and Dan Klein. 2008.
\newblock \href {https://www.aclweb.org/anthology/P08-1088} {Learning bilingual
  lexicons from monolingual corpora}.
\newblock In \emph{Proceedings of ACL-08: HLT}, pages 771--779, Columbus, Ohio.
  Association for Computational Linguistics.

\bibitem[{Hartmann et~al.(2019)Hartmann, Kementchedjhieva, and
  S{\o}gaard}]{hartmann_NIPS2019_8836}
Mareike Hartmann, Yova Kementchedjhieva, and Anders S{\o}gaard. 2019.
\newblock \href
  {http://papers.nips.cc/paper/8836-comparing-unsupervised-word-translation-methods-step-by-step.pdf}
  {Comparing unsupervised word translation methods step by step}.
\newblock In \emph{Advances in Neural Information Processing Systems 32}, pages
  6033--6043.

\bibitem[{Heafield(2011)}]{heafield-2011-kenlm}
Kenneth Heafield. 2011.
\newblock \href {https://www.aclweb.org/anthology/W11-2123} {{K}en{LM}: Faster
  and smaller language model queries}.
\newblock In \emph{Proceedings of the Sixth Workshop on Statistical Machine
  Translation}, pages 187--197, Edinburgh, Scotland. Association for
  Computational Linguistics.

\bibitem[{Hoshen and Wolf(2018)}]{hoshen-wolf-2018-non}
Yedid Hoshen and Lior Wolf. 2018.
\newblock \href {https://doi.org/10.18653/v1/D18-1043} {Non-adversarial
  unsupervised word translation}.
\newblock In \emph{Proceedings of the 2018 Conference on Empirical Methods in
  Natural Language Processing}, pages 469--478, Brussels, Belgium. Association
  for Computational Linguistics.

\bibitem[{Kalchbrenner and Blunsom(2013)}]{kalchbrenner2013recurrent}
Nal Kalchbrenner and Phil Blunsom. 2013.
\newblock \href {https://www.aclweb.org/anthology/D13-1176} {Recurrent
  continuous translation models}.
\newblock In \emph{Proceedings of the 2013 Conference on Empirical Methods in
  Natural Language Processing}, pages 1700--1709, Seattle, Washington, USA.
  Association for Computational Linguistics.

\bibitem[{Kim et~al.(2020)Kim, Gra\c{c}a, and Ney}]{kim-unsup-useless-2020}
Y.~Kim, M.~Gra\c{c}a, and H.~Ney. 2020.
\newblock \href {http://arxiv.org/abs/arXiv:2004.10581} {When and why is
  unsupervised neural machine translation useless?}
\newblock \emph{arXiv:2004.10581}.

\bibitem[{Kingma and Ba(2015)}]{kingma2014adam}
Diederik~P. Kingma and Jimmy Ba. 2015.
\newblock \href {http://arxiv.org/abs/1412.6980} {Adam: {A} method for
  stochastic optimization}.
\newblock In \emph{3rd International Conference on Learning Representations,
  {ICLR} 2015, San Diego, CA, USA, May 7-9, 2015, Conference Track
  Proceedings}.

\bibitem[{Koehn et~al.(2007)Koehn, Hoang, Birch, Callison-Burch, Federico,
  Bertoldi, Cowan, Shen, Moran, Zens, Dyer, Bojar, Constantin, and
  Herbst}]{koehn-etal-2007-moses}
Philipp Koehn, Hieu Hoang, Alexandra Birch, Chris Callison-Burch, Marcello
  Federico, Nicola Bertoldi, Brooke Cowan, Wade Shen, Christine Moran, Richard
  Zens, Chris Dyer, Ond{\v{r}}ej Bojar, Alexandra Constantin, and Evan Herbst.
  2007.
\newblock \href {https://www.aclweb.org/anthology/P07-2045} {{M}oses: Open
  source toolkit for statistical machine translation}.
\newblock In \emph{Proceedings of the 45th Annual Meeting of the Association
  for Computational Linguistics Companion Volume Proceedings of the Demo and
  Poster Sessions}, pages 177--180, Prague, Czech Republic. Association for
  Computational Linguistics.

\bibitem[{Koehn and Knight(2000)}]{koehn-knight-2000}
Philipp Koehn and Kevin Knight. 2000.
\newblock \href {https://www.aaai.org/Papers/AAAI/2000/AAAI00-109.pdf}
  {Estimating word translation probabilities from unrelated monolingual corpora
  using the em algorithm}.
\newblock In \emph{Proceedings of the Seventeenth National Conference on
  Artificial Intelligence and Twelfth Conference on Innovative Applications of
  Artificial Intelligence}, page 711–715. AAAI Press.

\bibitem[{Koehn and Knight(2002)}]{koehn-knight-2002-learning}
Philipp Koehn and Kevin Knight. 2002.
\newblock \href {https://doi.org/10.3115/1118627.1118629} {Learning a
  translation lexicon from monolingual corpora}.
\newblock In \emph{Proceedings of the {ACL}-02 Workshop on Unsupervised Lexical
  Acquisition}, pages 9--16, Philadelphia, Pennsylvania, USA. Association for
  Computational Linguistics.

\bibitem[{Lample et~al.(2018{\natexlab{a}})Lample, Conneau, Denoyer, and
  Ranzato}]{DBLP:conf/iclr/LampleCDR18}
Guillaume Lample, Alexis Conneau, Ludovic Denoyer, and Marc'Aurelio Ranzato.
  2018{\natexlab{a}}.
\newblock \href {https://openreview.net/forum?id=rkYTTf-AZ} {Unsupervised
  machine translation using monolingual corpora only}.
\newblock In \emph{6th International Conference on Learning Representations,
  {ICLR} 2018, Vancouver, BC, Canada, April 30 - May 3, 2018, Conference Track
  Proceedings}. OpenReview.net.

\bibitem[{Lample et~al.(2018{\natexlab{b}})Lample, Ott, Conneau, Denoyer, and
  Ranzato}]{lample-phrase-2018}
Guillaume Lample, Myle Ott, Alexis Conneau, Ludovic Denoyer, and Marc'Aurelio
  Ranzato. 2018{\natexlab{b}}.
\newblock \href {http://arxiv.org/abs/1804.07755} {Phrase-based {\&} neural
  unsupervised machine translation}.
\newblock \emph{CoRR}, abs/1804.07755.

\bibitem[{Liu et~al.(2020)Liu, Gu, Goyal, Li, Edunov, Ghazvininejad, Lewis, and
  Zettlemoyer}]{liu2020multilingual}
Yinhan Liu, Jiatao Gu, Naman Goyal, Xian Li, Sergey Edunov, Marjan
  Ghazvininejad, Mike Lewis, and Luke Zettlemoyer. 2020.
\newblock Multilingual denoising pre-training for neural machine translation.
\newblock \emph{arXiv preprint arXiv:2001.08210}.

\bibitem[{Mikolov et~al.(2013{\natexlab{a}})Mikolov, Le, and
  Sutskever}]{mikolov2013exploiting}
Tomas Mikolov, Quoc~V Le, and Ilya Sutskever. 2013{\natexlab{a}}.
\newblock \href {https://arxiv.org/abs/1309.4168} {Exploiting similarities
  among languages for machine translation}.
\newblock \emph{arXiv preprint arXiv:1309.4168}.

\bibitem[{Mikolov et~al.(2013{\natexlab{b}})Mikolov, Sutskever, Chen, Corrado,
  and Dean}]{mikolov2013}
Tomas Mikolov, Ilya Sutskever, Kai Chen, Greg~S Corrado, and Jeff Dean.
  2013{\natexlab{b}}.
\newblock Distributed representations of words and phrases and their
  compositionality.
\newblock In \emph{Advances in neural information processing systems}, pages
  3111--3119.

\bibitem[{Nakashole and Flauger(2018)}]{nakashole-flauger-2018-characterizing}
Ndapa Nakashole and Raphael Flauger. 2018.
\newblock \href {https://doi.org/10.18653/v1/P18-2036} {Characterizing
  departures from linearity in word translation}.
\newblock In \emph{Proceedings of the 56th Annual Meeting of the Association
  for Computational Linguistics (Volume 2: Short Papers)}, pages 221--227,
  Melbourne, Australia. Association for Computational Linguistics.

\bibitem[{Och(2003)}]{och-2003-minimum}
Franz~Josef Och. 2003.
\newblock \href {https://doi.org/10.3115/1075096.1075117} {Minimum error rate
  training in statistical machine translation}.
\newblock In \emph{Proceedings of the 41st Annual Meeting of the Association
  for Computational Linguistics}, pages 160--167, Sapporo, Japan. Association
  for Computational Linguistics.

\bibitem[{Ormazabal et~al.(2019)Ormazabal, Artetxe, Labaka, Soroa, and
  Agirre}]{ormazabal-etal-2019-analyzing}
Aitor Ormazabal, Mikel Artetxe, Gorka Labaka, Aitor Soroa, and Eneko Agirre.
  2019.
\newblock \href {https://doi.org/10.18653/v1/P19-1492} {Analyzing the
  limitations of cross-lingual word embedding mappings}.
\newblock In \emph{Proceedings of the 57th Annual Meeting of the Association
  for Computational Linguistics}, pages 4990--4995, Florence, Italy.
  Association for Computational Linguistics.

\bibitem[{Ott et~al.(2019)Ott, Edunov, Baevski, Fan, Gross, Ng, Grangier, and
  Auli}]{fairseq}
Myle Ott, Sergey Edunov, Alexei Baevski, Angela Fan, Sam Gross, Nathan Ng,
  David Grangier, and Michael Auli. 2019.
\newblock \href {https://doi.org/10.18653/v1/N19-4009} {fairseq: A fast,
  extensible toolkit for sequence modeling}.
\newblock In \emph{Proceedings of the 2019 Conference of the North {A}merican
  Chapter of the Association for Computational Linguistics (Demonstrations)},
  pages 48--53, Minneapolis, Minnesota. Association for Computational
  Linguistics.

\bibitem[{Papineni et~al.(2002)Papineni, Roukos, Ward, and
  Zhu}]{papineni2002bleu}
Kishore Papineni, Salim Roukos, Todd Ward, and Wei-Jing Zhu. 2002.
\newblock Bleu: a method for automatic evaluation of machine translation.
\newblock In \emph{Proceedings of the 40th annual meeting on association for
  computational linguistics}, pages 311--318. Association for Computational
  Linguistics.

\bibitem[{Patra et~al.(2019)Patra, Moniz, Garg, Gormley, and
  Neubig}]{patra-etal-2019-bilingual}
Barun Patra, Joel Ruben~Antony Moniz, Sarthak Garg, Matthew~R. Gormley, and
  Graham Neubig. 2019.
\newblock \href {https://doi.org/10.18653/v1/P19-1018} {Bilingual lexicon
  induction with semi-supervision in non-isometric embedding spaces}.
\newblock In \emph{Proceedings of the 57th Annual Meeting of the Association
  for Computational Linguistics}, pages 184--193, Florence, Italy. Association
  for Computational Linguistics.

\bibitem[{Post(2018)}]{post-2018-call}
Matt Post. 2018.
\newblock \href {https://www.aclweb.org/anthology/W18-6319} {A call for clarity
  in reporting {BLEU} scores}.
\newblock In \emph{Proceedings of the Third Conference on Machine Translation:
  Research Papers}, pages 186--191, Belgium, Brussels. Association for
  Computational Linguistics.

\bibitem[{Rapp(1995)}]{rapp1995}
Reinhard Rapp. 1995.
\newblock \href {https://doi.org/10.3115/981658.981709} {Identifying word
  translations in non-parallel texts}.
\newblock In \emph{Proceedings of the 33rd Annual Meeting on Association for
  Computational Linguistics}, ACL ’95, page 320–322, USA. Association for
  Computational Linguistics.

\bibitem[{Ravi and Knight(2011)}]{ravi2011deciphering}
Sujith Ravi and Kevin Knight. 2011.
\newblock \href {https://www.aclweb.org/anthology/P11-1002} {Deciphering
  foreign language}.
\newblock In \emph{Proceedings of the 49th Annual Meeting of the Association
  for Computational Linguistics: Human Language Technologies}, pages 12--21,
  Portland, Oregon, USA. Association for Computational Linguistics.

\bibitem[{Ruder et~al.(2019)Ruder, Vuli{\'c}, and S{\o}gaard}]{ruder2019survey}
Sebastian Ruder, Ivan Vuli{\'c}, and Anders S{\o}gaard. 2019.
\newblock \href {https://doi.org/10.1613/jair.1.11640} {A survey of
  cross-lingual word embedding models}.
\newblock \emph{Journal of Artificial Intelligence Research}, 65:569--631.

\bibitem[{Sennrich et~al.(2016{\natexlab{a}})Sennrich, Haddow, and
  Birch}]{sennrich2016improving}
Rico Sennrich, Barry Haddow, and Alexandra Birch. 2016{\natexlab{a}}.
\newblock \href {https://doi.org/10.18653/v1/P16-1009} {Improving neural
  machine translation models with monolingual data}.
\newblock In \emph{Proceedings of the 54th Annual Meeting of the Association
  for Computational Linguistics (Volume 1: Long Papers)}, pages 86--96, Berlin,
  Germany. Association for Computational Linguistics.

\bibitem[{Sennrich et~al.(2016{\natexlab{b}})Sennrich, Haddow, and
  Birch}]{sennrich-etal-2016-neural}
Rico Sennrich, Barry Haddow, and Alexandra Birch. 2016{\natexlab{b}}.
\newblock \href {https://doi.org/10.18653/v1/P16-1162} {Neural machine
  translation of rare words with subword units}.
\newblock In \emph{Proceedings of the 54th Annual Meeting of the Association
  for Computational Linguistics (Volume 1: Long Papers)}, pages 1715--1725,
  Berlin, Germany. Association for Computational Linguistics.

\bibitem[{S{\o}gaard et~al.(2018)S{\o}gaard, Ruder, and
  Vuli{\'c}}]{sogaard-etal-2018-limitations}
Anders S{\o}gaard, Sebastian Ruder, and Ivan Vuli{\'c}. 2018.
\newblock \href {https://doi.org/10.18653/v1/P18-1072} {On the limitations of
  unsupervised bilingual dictionary induction}.
\newblock In \emph{Proceedings of the 56th Annual Meeting of the Association
  for Computational Linguistics (Volume 1: Long Papers)}, pages 778--788,
  Melbourne, Australia. Association for Computational Linguistics.

\bibitem[{Song et~al.(2019)Song, Tan, Qin, Lu, and Liu}]{song2019mass}
Kaitao Song, Xu~Tan, Tao Qin, Jianfeng Lu, and Tie-Yan Liu. 2019.
\newblock Mass: Masked sequence to sequence pre-training for language
  generation.
\newblock In \emph{ICML}.

\bibitem[{Srivastava et~al.(2014)Srivastava, Hinton, Krizhevsky, Sutskever, and
  Salakhutdinov}]{srivastava2014dropout}
Nitish Srivastava, Geoffrey Hinton, Alex Krizhevsky, Ilya Sutskever, and Ruslan
  Salakhutdinov. 2014.
\newblock \href {http://jmlr.org/papers/v15/srivastava14a.html} {Dropout: A
  simple way to prevent neural networks from overfitting}.
\newblock \emph{Journal of Machine Learning Research}, 15(56):1929--1958.

\bibitem[{Sutskever et~al.(2014)Sutskever, Vinyals, and
  Le}]{sutskever2014sequence}
Ilya Sutskever, Oriol Vinyals, and Quoc~V Le. 2014.
\newblock \href
  {https://papers.nips.cc/paper/5346-sequence-to-sequence-learning-with-neural-networks.pdf}
  {Sequence to sequence learning with neural networks}.
\newblock In \emph{Advances in neural information processing systems}, pages
  3104--3112.

\bibitem[{Tiedemann(2012)}]{tiedemann2012parallel}
J{\"o}rg Tiedemann. 2012.
\newblock \href
  {http://www.lrec-conf.org/proceedings/lrec2012/pdf/463_Paper.pdf} {Parallel
  data, tools and interfaces in {OPUS}}.
\newblock In \emph{Proceedings of the Eighth International Conference on
  Language Resources and Evaluation ({LREC}'12)}, pages 2214--2218, Istanbul,
  Turkey. European Language Resources Association (ELRA).

\bibitem[{Vaswani et~al.(2017)Vaswani, Shazeer, Parmar, Uszkoreit, Jones,
  Gomez, Kaiser, and Polosukhin}]{vaswani_2017}
Ashish Vaswani, Noam Shazeer, Niki Parmar, Jakob Uszkoreit, Llion Jones,
  Aidan~N Gomez, \L~ukasz Kaiser, and Illia Polosukhin. 2017.
\newblock \href
  {http://papers.nips.cc/paper/7181-attention-is-all-you-need.pdf} {Attention
  is all you need}.
\newblock In I.~Guyon, U.~V. Luxburg, S.~Bengio, H.~Wallach, R.~Fergus,
  S.~Vishwanathan, and R.~Garnett, editors, \emph{Advances in Neural
  Information Processing Systems 30}, pages 5998--6008. Curran Associates, Inc.

\bibitem[{Vuli{\'c} et~al.(2019)Vuli{\'c}, Glava{\v{s}}, Reichart, and
  Korhonen}]{vulic-etal-2019-really}
Ivan Vuli{\'c}, Goran Glava{\v{s}}, Roi Reichart, and Anna Korhonen. 2019.
\newblock \href {https://doi.org/10.18653/v1/D19-1449} {Do we really need fully
  unsupervised cross-lingual embeddings?}
\newblock In \emph{Proceedings of the 2019 Conference on Empirical Methods in
  Natural Language Processing and the 9th International Joint Conference on
  Natural Language Processing (EMNLP-IJCNLP)}, pages 4407--4418, Hong Kong,
  China. Association for Computational Linguistics.

\bibitem[{Wo{\l}k and Marasek(2014)}]{wolk2014building}
Krzysztof Wo{\l}k and Krzysztof Marasek. 2014.
\newblock Building subject-aligned comparable corpora and mining it for truly
  parallel sentence pairs.
\newblock \emph{Procedia Technology}, 18:126--132.

\bibitem[{Yang et~al.(2018)Yang, Chen, Wang, and
  Xu}]{yang-etal-2018-unsupervised}
Zhen Yang, Wei Chen, Feng Wang, and Bo~Xu. 2018.
\newblock \href {https://doi.org/10.18653/v1/P18-1005} {Unsupervised neural
  machine translation with weight sharing}.
\newblock In \emph{Proceedings of the 56th Annual Meeting of the Association
  for Computational Linguistics (Volume 1: Long Papers)}, pages 46--55,
  Melbourne, Australia. Association for Computational Linguistics.

\bibitem[{Ziemski et~al.(2016)Ziemski, Junczys-Dowmunt, and
  Pouliquen}]{ziemski-etal-2016-united}
Micha{\l} Ziemski, Marcin Junczys-Dowmunt, and Bruno Pouliquen. 2016.
\newblock \href {https://www.aclweb.org/anthology/L16-1561} {The united nations
  parallel corpus v1.0}.
\newblock In \emph{Proceedings of the Tenth International Conference on
  Language Resources and Evaluation ({LREC}'16)}, pages 3530--3534,
  Portoro{\v{z}}, Slovenia. European Language Resources Association (ELRA).

\end{thebibliography}

\end{document}